\documentclass[runningheads]{llncs}
\usepackage{graphicx}
\usepackage{caption, subcaption} 

\begin{document}
\title{Face Recognition Using $Sf_{3}CNN$ With Higher Feature Discrimination}
%
%
\author{Nayaneesh Kumar Mishra\inst{1}\orcidID{0000-0002-2164-9541} \and
Satish Kumar Singh \inst{2}
}
\authorrunning{N. K. Mishra and S. K. Singh}
%
\institute{Indian Institute of Information Technology, Allahabad, India\\
\email{nayaneesh@gmail.com}\\
\and
Indian Institute of Information Technology, Allahabad, India\\
\email{satish@iiita.ac.in}}
\maketitle              
\begin{abstract}
With the advent of 2-dimensional Convolution Neural Networks (2D CNNs), the face recognition accuracy has reached above 99\%. However, face recognition is still a challenge in real world conditions. A video, instead of an image, as an input can be more useful to solve the challenges of face recognition in real world conditions. This is because a video provides more features than an image. However, 2D CNNs cannot take advantage of the temporal features present in the video. We therefore, propose a framework called $Sf_{3}CNN$ for face recognition in videos. The $Sf_{3}CNN$ framework uses 3-dimensional Residual Network (3D Resnet) and A-Softmax loss for face recognition in videos. The use of 3D ResNet helps to capture both spatial and temporal features into one compact feature map. However, the 3D CNN features must be highly discriminative for efficient face recognition. The use of A-Softmax loss helps to extract highly discriminative features from the video for face recognition. $Sf_{3}CNN$ framework gives an increased accuracy of 99.10\% on CVBL video database in comparison to the previous 97\% on the same database using 3D ResNets.

\end{abstract}

\section{Introduction}

With the advent of deep learning, 2-dimensional Convolution Neural Networks (2D CNN) came to be used for recognition of faces \cite{imagenet,deep_1,deepface,deepid3,pose_invariant_fr,l2_norm}. The accuracy of 2D CNN architectures for face recognition has reached to 99.99\% \cite{facenet,deep_2} as reported on LFW database. In-spite of these near-hundred-percent accuracies, the face recognition algorithms fail when applied in real world conditions because of the real world challenges such as varying pose, illumination, occlusion and resolution. In this paper, we propose to overcome these limitations of the face recognition algorithms by the use of 3-dimensional Convolution Neural Networks (3D CNN). This is because a 3D CNN processes video as an input and extracts both temporal as well as spatial information from the video into a compact feature. The compact feature generated from 3D CNN contains more information than that generated using 2D CNN. This allows for an efficient and robust face recognition \cite{3dcnn}. The features generated using 3D CNNs however, are not highly discriminative and hence affect the accuracy \cite{3dcnn}. The discriminative ability of the loss functions has been increased by using the concept of angular margin. A-softmax \cite{sphereface} is the most basic loss in the series of the loss functions that implements the concept of angular margin in the most naive form. Hence we propose to use A-softmax for face recognition with 3D CNN. 

This paper, therefore, has the following contribution: 

We develop a deep learning framework called $Sf_{3}CNN$ that uses 3D CNN and A-softmax loss for efficient and robust face recognition in videos.

\section{Proposed Architecture} \label{section:proposed_architecture}
\begin{figure}[!t]
\centering
    \includegraphics[scale=.45]{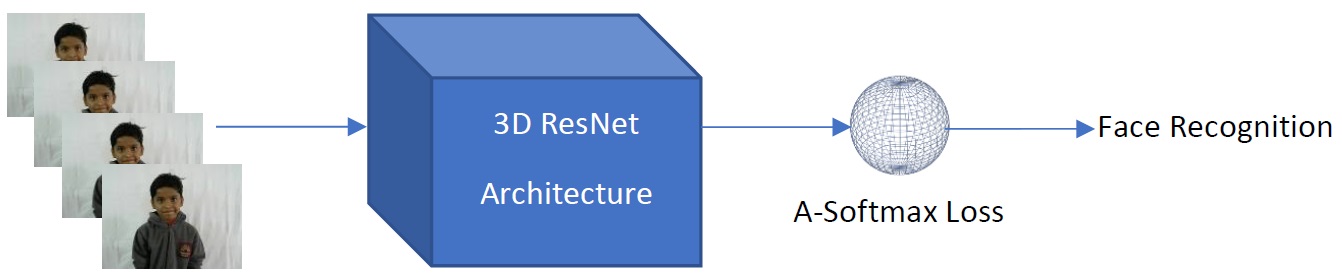}
\caption{$Sf_{3}CNN$ is a very simple framework that contains 3D Residual Network followed by Angular Softmax Loss for better discrimination among the class features. $Sf_{3}CNN$ framework is experimented by substituting the 3D Resnet Architecture block in the figure by variants of 3D-Resnet architectures. $Sf_{3}CNN$ framework gave the best accuracy of 99.10\% on CVBL video database by using Resnet-152 and Wideresnet-50. Finally, Resnet-152 is chosen as the 3D-Resnet architecture for the framework because of comparatively less number of parameters in comparison to Wideresnet-50.}
\label{fig:Sf3CNN_Architecture}
\end{figure}

We propose a 3D CNN framework for face recognition in videos. The framework is called $Sf_{3}CNN$ as shown in figure \ref{fig:Sf3CNN_Architecture}. The $Sf_{3}CNN$ framework uses 3D ResNets for feature extraction from the input video followed by A-softmax loss. The use of 3D CNN helps to extract compact features from the video which contains both spatial and temporal information. A-softmax loss achieves high feature discrimination. The $Sf_{3}CNN$ framework is named so because the '$Sf$' in the name represents the A-Softmax loss which is a variant of the Softmax loss and the term '$_{3}CNN$' represents the 3D CNN in the framework.


\section{A-softmax loss}

A-softmax loss \cite{sphereface} is developed from Softmax loss to increase the discriminative ability of Softmax loss function. A-softmax loss introduces angular margin in the softmax loss to maximize the inter-class distance and minimise the intra-class distance among the class features.

For  an input feature $x_{i}$ with label $y_{i}$, angle $\theta_{j, i}$ is the angle between the input vector $x_{i}$ and weight vector $W_{j}$  for any class $j$. When we normalize $||W_{j}|| = 1 $ for all $j$  and zero the biases and also increase the angular margin between the features of the actual class and the the rest of the features by multiplying the angle $\theta_{y_{i}, i}$ with $m$, we get the equation for A-softmax as:
\begin{equation}
\label{equation:sphereface}
    L_{ang} = \frac{1}{N} \sum_{i} -log \left ( \frac{e^{\left \| x_{i} \right \|\cos(m\theta_{y_{i}, i})}}{e^{\left \| x_{i} \right \|\cos(m\theta_{y_{i}, i})} +  \sum_{j \neq y_{i}} e^{\left \| x_{i} \right \|\cos(\theta_{j, i})}
} \right )
\end{equation}
In equation \ref{equation:sphereface}, $L_{ang}$ is called the A-softmax loss. N is the number of training samples over which the mean of the loss is calculated. $\theta_{y_{i}, i}$ has to be in the range of $[0, \pi/m]$. From equation \ref{equation:sphereface}, it is clear that A-softmax loss \cite{sphereface} increases the angular margin between the feature vector of the actual class and the input vector by introducing the hyperparameter $m$. It thus tries to maximize the inter-class cosine distance and minimize the intra-class cosine distance among the features. The angular margin in A-softmax loss thus increases the discriminative ability of the loss function. 

\section{Implementation} \label{section:implementation}

We performed the experiments on CVBL (Computer Vision and Biometric Lab) database \cite{3dcnn} using $Sf_{3}CNN$ framework. The optimizer used is Adamax. The initial learning rate has been kept to 0.002. The values of the parameter betas is equal to 0.9 and 0.999. The value of the parameter eps is 1e-08  and weight decay is zero. The activation function used is Parametric Rectified Linear Unit (PReLU) because PReLU has been found to work better with A-softmax loss \cite{sphereface}.

We have compared our results with the work on face recognition by Mishra et. al. \cite{3dcnn}. In the work by Mishra et. al. \cite{3dcnn}, the experiment has been performed on CVBL database using 3D residual networks when the loss function is Cross-entropy. Both, in our experiment and in the experiment performed by Mishra et. al. \cite{3dcnn}, a clip of temporal length 16 is input to 3D residual network of different depths and genres. In case the number of frames are less than 16, the same frames are repeated by looping around the existing frames. Horizontal flipping is done on frames with probability of 50 percent. Cropping is performed from one of the locations out of 4 corners and 1 center in the original frame. This is then scaled on the basis of the a scale value selected out of $\left\{ \frac{1}{2^{\frac{1}{4}}} ,  \frac{1}{\sqrt{2}} ,  \frac{1}{2^{\frac{3}{4}}},  \frac{1}{2} \right\}$. The scaling is done by maintaining the aspect ratio to one and selecting the shorter length of the image. All the scaled frames are re-sized to 112 by 112 for input to the architecture. Mean subtraction is performed on each channel by subtraction of a mean value from each channel. This is done to keep the pixel values zero centered. The CVBL video database is divided into 60:40 ratio for training and validation purpose. Out of the total 675 videos, 415 videos have been taken for training and the rest 260 videos have been considered for validation.

\section{Results and Discussion}
\begin{figure}
\begin{subfigure}{0.5\textwidth}
\includegraphics[width=0.9\linewidth, height=5cm]{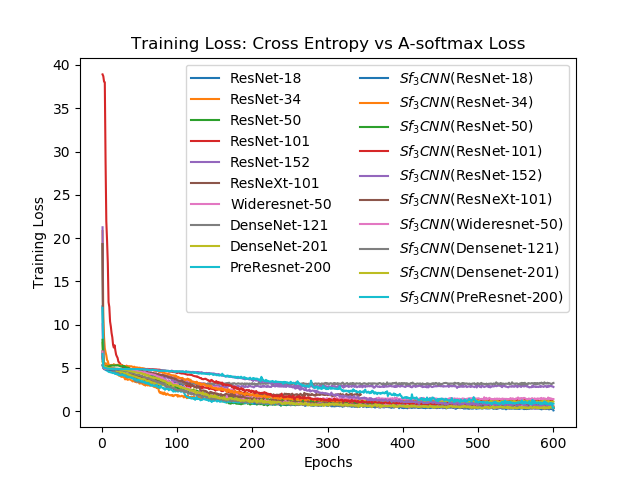} 
\caption{ResNet-50}
\label{resnet-50_train_val_loss}
\end{subfigure}
\begin{subfigure}{0.5\textwidth}
\includegraphics[width=0.9\linewidth, height=5cm]{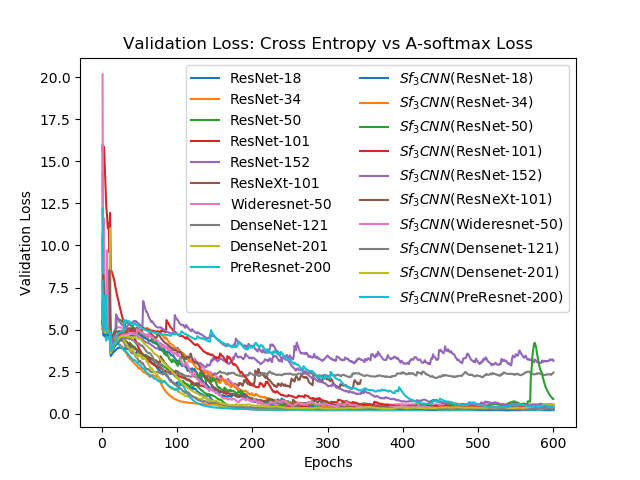} 
\caption{Densenet-121}
\label{densenet-121_train_val_loss}
\end{subfigure}
\caption{Comparison of $Sf_{3}CNN$ Framework with 3D CNN + Cross Entropy Loss. (a) Comparison based on Training Loss (b) Comparison based on Validation Loss (c) Comparison based on Validation Accuracy}
\label{fig:comparison_cross_entropy_sphereface}
\end{figure}

The results of $Sf_{3}CNN$ framework of various depths and genres on CVBL database are summarized Table \ref{accuracy_table}. $Sf_{3}CNN$ framework uses activation function PReLU and optimization function Adamax. PReLU is used with A-softmax in our experiment because PReLU is recommended to be used with A-softmax \cite{sphereface}. For the purpose of comparison, Table \ref{accuracy_table} also shows results obtained by Mishra et. al. \cite{3dcnn}. In the work by Mishra et. al. \cite{3dcnn}, face recognition is done using 3D CNN with cross-entropy loss, SGD (Stochastic Gradient Descent) optimizer and ReLU (Rectified Linear Unit) activation function. Graph in Figure \ref{fig:comparison_cross_entropy_sphereface} shows the comparison between the performance of the $Sf_{3}CNN$ framework and the architecture used by Mishra et. al. \cite{3dcnn} in terms of how the training loss and validation loss varies with the number of epochs.

\begin{table}
\caption{Comparison of validation accuracy of $Sf_{3}CNN$ Framework with state-of-art methods on CVBL database for face recognition}
\label{accuracy_table}
\begin{tabular}{|l|c|c|}
\hline
{\bfseries Residual Networks} & {\bfseries Accuracy for }            & {\bfseries Accuracy for }       \\
                              & {\bfseries 3D CNN +     }            & { \bfseries $Sf_{3}CNN$ }      \\ 
                              & {\bfseries Cross Entropy Loss in \%} & { \bfseries Framework in \% }     \\
\hline
ResNet-18    &  96  &  98.97 \\
ResNet-34    &  93.7 &  98.72 \\
ResNet-50    &  96.2 &  98.59 \\
ResNet-101   &  93.4 &  98.72 \\
ResNet-152   &  49.1 & \textbf{99.10} \\
ResNeXt-101  &  78.5 & 98.59 \\
Pre-activation ResNet-200 & 96.2  & 98.46  \\
Densenet-121 & 55  & 98.72  \\
Densenet-201 & \textbf{97}  & 98.33  \\
WideResnet-50& 90.2   & \textbf{99.10}  \\
\hline
\end{tabular}
\end{table}

In Figure \ref{fig:comparison_cross_entropy_sphereface}, convergence of loss and accuracy in case of training and validation for the $Sf_{3}CNN$ framework is shown for different variants of 3D-Resnets in $Sf_{3}CNN$ framework. We have taken running average of the validation loss and validation accuracy to highlight the general path of convergence. We can easily observe that convergence of loss in case of training and validation for the $Sf_{3}CNN$ framework is almost same as obtained by Mishra et. al. \cite{3dcnn}. The training and the corresponding validation graph almost follow the same path and converge to nearly the same loss value, indicating that low accuracy was never because of underfitting and high accuracy was never because of overfitting. Thus we can conclude that both $Sf_{3}CNN$ and the architectures used by Mishra et. al. \cite{3dcnn} were stable throughout their training.

From the Table \ref{accuracy_table} it can be observed that, $Sf_{3}CNN$ framework achieves improvement in accuracy in face recognition for all depths and genres of 3D ResNets in comparison to the results obtained by Mishra et. al. \cite{3dcnn}. For $Sf_{3}CNN$ framework, the difference between the highest and lowest accuracy is just 0.77\% in comparison to 47.9\% in case of Mishra et. al. \cite{3dcnn}. For $Sf_{3}CNN$ framework, the accuracy varies between 98\% and 99.10\% and therefore does not significantly vary with depth. In case of Mishra et. al. \cite{3dcnn} however, the accuracy varies from 49.1\% to 97\%. Thus we can easily infer that unlike in case of Mishra et. al. \cite{3dcnn}, the $Sf_{3}CNN$ framework is successful in increasing the discrimination between classes, so much so, that it mitigates the role of depth of the architecture on the accuracy of face recognition.


$Sf_{3}CNN$ framework successfully achieves the highest accuracy of 99.10\% which is well above the the highest accuracy of 97\% as obtained by Mishra et. al. \cite{3dcnn}. $Sf_{3}CNN$ framework achieved the highest accuracy with ResNet-152 and Wideresnet-50. It is interesting to note that ResNet-152 had achieved the lowest accuracy of just 49.1\% and Wideresnet-50 had achieved an accuracy of just 90.2\% in the work by Mishra et. al. \cite{3dcnn}. It is just because of the highly discriminative nature of the A-softmax loss function in $Sf_{3}CNN$ framework that ResNet-152 could rise up to give the highest accuracy.
Because of similar reasons, the accuracy of Densenet-201, which was 97\% in the work by Mishra et. al. \cite{3dcnn}, rose to 98.33\% in case of $Sf_{3}CNN$ framework.

As can be seen in Table \ref{accuracy_table}, both ResNet-152 and Wideresnet-50 achieved the highest accuracy of 99.10\%. Since Wideresnet-50 contains far more number of parameters than in comparison to Resnet-152 \cite{kensho}, it would therefore be preferable to use Resnet-152 instead of Wideresnet-50 in the $Sf_{3}CNN$ framework. At the same time, it is also noticeable that there are comparatively more number of feature maps for each convolutional layer in Wideresnet-50. Wideresnet-50 is therefore efficient in parallel computing using GPUs (Graphics Processing Unit) \cite{kensho}. Hence, if the the number of GPUs  are more than one, Wideresnet-50 can be used in $Sf_{3}CNN$ framework to take advantage of parallel computing using GPUs.

\section{Conclusion}

In this paper, we proposed a framework called $Sf_{3}CNN$. Based on the experimentation results on the CVBL video database, it can be concluded that $Sf_{3}CNN$ framework is capable of robust face recognition even in real world conditions. The high discriminative ability of $Sf_{3}CNN$ framework leads to increased accuracy of face recognition to 99.10\% which is better than the highest accuracy of 97\% in the work by Mishra et. al. \cite{3dcnn}.

%
%
%
 \bibliographystyle{splncs04}
 \bibliography{mybibliography}
\end{document}